\documentclass[nai,crcready]{iosart2x}

\usepackage{subcaption}

\pubyear{0000}
\volume{0}
\firstpage{1}
\lastpage{1}

\begin{document}

\begin{frontmatter}

\title{Machine Learning with Requirements: a Manifesto}
\runtitle{Machine Learning with Requirements: a Manifesto}

\begin{aug}
\author[A]{\inits{E.}\fnms{Eleonora} \snm{Giunchiglia}\ead[label=e1]{eleonora.giunchiglia@tuwien.ac.at}%
\thanks{Corresponding author. \printead{e1}.}}
\author[B]{\inits{F.}\fnms{Fergus} \snm{Imrie}\ead[label=e2]{imrie@ucla.edu}}
\author[C,E]{\inits{M.}\fnms{Mihaela} \snm{van der Schaar}\ead[label=e3]{mv472@cam.ac.uk}}
\author[A,D]{\inits{T.}\fnms{Thomas} \snm{Lukasiewicz}\ead[label=e4]{thomas.lukasiewicz@tuwien.ac.at}}
\address[A]{Institute of Logic and Computation, \orgname{TU Wien},
 \cny{Austria}\printead[presep={\\}]{e1,e4}}
  \address[B]{Department of Electrical and Computer Engineering, \orgname{University of California, Los Angeles}, \cny{ USA}\printead[presep={\\}]{e2}}
  \address[C]{DAMTP, \orgname{University of Cambridge},
 \cny{United Kingdom}\printead[presep={\\}]{e3}}
\address[D]{Department of Computer Science, \orgname{University of Oxford},
 \cny{United Kingdom}}%
 \address[E]{\orgname{Alan Turing Institute},
 \cny{United Kingdom}}
\end{aug}

\begin{abstract}
In the recent years, machine learning has made great advancements that have been at the root of many breakthroughs in different application domains. However, it is still an open issue how make them applicable to  high-stakes or safety-critical application domains, as they can often be brittle and unreliable.
In this paper, we argue that requirements definition and satisfaction can go a long way to make machine learning models 
even more fitting to the real world, especially  in critical domains. 
To this end, we present two problems in which $(i)$ requirements arise naturally, $(ii)$ machine learning models are or can be fruitfully deployed, and $(iii)$ neglecting the requirements can have dramatic consequences.
We show how the requirements specification  can be fruitfully integrated into the standard machine learning development pipeline,  proposing a novel pyramid development process in which requirements definition may impact all the subsequent phases in the pipeline, and viceversa.

\end{abstract}

\begin{keyword}
\kwd{Machine Learning}
\kwd{Requirements}
\kwd{Machine Learning Operations}
\kwd{Software Engineering}
\end{keyword}

\end{frontmatter}

\section{Introduction}\label{sec:intro}

In the recent years, machine learning has made great advancements that have been at the root of many breakthroughs in
different application domains. For example, AlphaFold~\cite{senior2020alphafold} is a deep learning model that solved the ``protein folding problem'', a grand challenge in the field of biology for more than half a century, while  halicin~\cite{stokes2020antibiotic} is the first antibiotic discovered using machine learning, that could help in the battle against bacterial resistance. Results like those above, though ground-breaking, overshadow the dangers that come with the careless use of machine learning models in critical applications. Indeed, even though such models report astonishingly high-performance in terms of accuracy---or alternatively chosen metric---they do not give
any guarantee that the model will not have any unintended behaviour when used in practice. Indeed, as pointed out by Rudin in \cite{rudin2019nature}, there have already been cases of people erroneously denied parole~\cite{wexler2017}, deep learning models monitoring the air quality in the US that claimed that highly polluted air could be safely breathed~\cite{mcgough2018} and generally poor use of limited resources in medicine, criminal justice, finance and in other domains~\cite{ramazon2016}. Such problems in the behavior of the model are often rooted in the quality of the dataset used for training the model. This is for example the case
happened in the early stages of the the Covid-19 pandemic, where, as reported in \cite{roberts2021_covid_problems}, a common chest scan dataset from~\cite{kermany2018_chest_child_dataset} made of pediatric scans, was used as control group against Covid-19 positive scans: as a result, instead of Covid-19 detectors, adult/child classifiers were built.
Independently from the reasons, such unintended outcomes $(i)$ are particularly dangerous in safety-critical applications, where even a single unforeseen mistake can lead to dramatic consequences, and $(ii)$ undermine the human confidence in the models themselves, thus slowing down their adoption. 

In this paper, we argue that requirements
specification and verification can go a long way to make machine learning models even more fitting to the real world, by reducing the risk of getting potentially dangerous unintended behaviors. We start from the simple observation  that unintended behaviors correspond to the model violating  some requirements which may be known even before the data collection and model development start. To support this claim, we consider two examples, one in healthcare the other in autonomous driving, in which $(i)$ some requirements are known in advance, $(ii)$ machine learning models have very positive performance in terms of accuracy (or other selected metric), and $(iii)$ despite the positive performance, the outcome of standardly developed machine learning models often violates the spelled requirements with possible dramatic consequences given the criticality of the application domains. Though we consider just two examples, we believe that analogous considerations apply to most application domains given 
$(i)$ the 
 body of knowledge developed along the years in any application domain, which can be translated in corresponding requirements, $(ii)$ the impressive results in performance obtained by machine learning models, which will continue to push for their adoption despite the possible unintended behaviors, and $(iii)$ the fact that it is not surprising the resulting model will violate the requirements if they are not taken into account during the data-collection and model development. 
 Refining the last observation, it is not surprising a machine learning model has an unintended behavior if, before its deployment, the data and the model itself are not somehow verified against the  requirement capturing the intended behavior.
We therefore claim that the requirements' definition should precede and involve the entire machine learning model development cycle, including the dataset construction. In particular, we propose a novel {\sl pyramid} machine learning development process which  $(i)$  highlights how the requirement definition can help improving every single step of the standard machine learning model development process, and $(ii)$ takes into account that some requirement adjustment might be necessary because of difficulties and/or discoveries that emerge during the development process of machine learning models, especially given their data-driven  nature.  
The above proposal can be seen as a call to adapt and adopt the general methodologies used in software engineering for generic system development, where it is well known  $(i)$ that the later the requirement are taken into account the higher the cost is to repair the system, and also $(ii)$ that the requirement definition is an iterative process in which requirements affect the system development and viceversa, given that during the system development some requirement may be newly discovered and/or adjusted if not canceled, see, e.g., \cite{sommerville,sommerville2007_soft_eng,nuseibeh}. We will ground the above facts in the specific machine learning model development pipeline, highlighting the pros 
of adopting requirement in the different phases.

Our proposal represents a significant shift from the traditional {\sl ``performance-driven''} machine learning development pipeline, which solely concentrates on how to get better performance, measured in terms of accuracy or alternatively chosen metric. Indeed, we are proposing a {\sl ``requirements-driven''} machine learning development pipeline in which performance is just one of the many requirements the model has to satisfy. Our proposal is thus in line with all the mostly recent works which do not only focus on performance, but also check that the model satisfies other properties like fairness, robustness, explainability, sustainability and safety. This obviously might make the development process more complex, because different requirements might be contradictory with one another: a problem that is well known in software engineering~\cite{sommerville}. 
For example, at an high level of abstraction, complex models will likely have better performance at the possible price of being less explainable and/or verifiable and/or sustainable. Analogously, it might be the case that satisfying certain fairness properties (e.g., equalized odds or demographic parity) is preferable to a higher performance. Independently from the properties we wish for the model, it is clear that $(i)$ the sooner the requirement are made explicit the better, and $(ii)$ taking into account the requirement in the dataset definition and model construction will likely lead to model which will be easier to verify with respect the stated requirements. 
To this end, a good example is given by the recent research in the neuro-symbolic field, where researchers have started developing  models that are guaranteed by-design to be compliant with a set of given requirements expressed as logical constraints (see e.g., \cite{giunchiglia2021,hoernle2022multiplexnet,giunchiglia2022_road}): in these works it has been shown how the requirements can be incorporated in the model which automatically satisfy them, and have a positive impact also on performance.

Our proposal enjoys the same spirit of  works such as \cite{gebru2018_datasheets_for_datasets,mitchell2019,mihaela2022-dc-check} which advocate for a more structured approach to machine learning model development. In particular, Gebru et al. in \cite{gebru2018_datasheets_for_datasets} propose documentation guidelines for new datasets, Mitchell et al. in \cite{mitchell2019} advocate for standardized reporting of models, including training data, performance measures, and limitations, while Seedat et al. \cite{mihaela2022-dc-check} propose a actionable checklist-style framework to elicit data-centric considerations at different stages of the development pipeline.

The remaining of the paper is structured as follows. 
First, in Section~\ref{sec:need}, we consider examples from the healthcare and autonomous driving application domains in order to show  that in these domains  $(i)$ requirements arise naturally,  $(ii)$ machine learning models are or can be fruitfully
deployed, and $(iii)$ neglecting the requirements can have dramatic consequences. Secondly, in Section~\ref{sec:pyramid}, we show how the requirements definition  can be fruitfully integrated into the standard machine learning development pipeline,  impacting all the phases in the pipeline.
Finally, we have the conclusions and possible plans for the road ahead in Section~\ref{sec:concl}. 

\section{The Need for Requirements}\label{sec:need}

\begin{figure}[t]
\includegraphics[width=.8\textwidth,trim={0cm 5cm 0cm 5cm},clip]{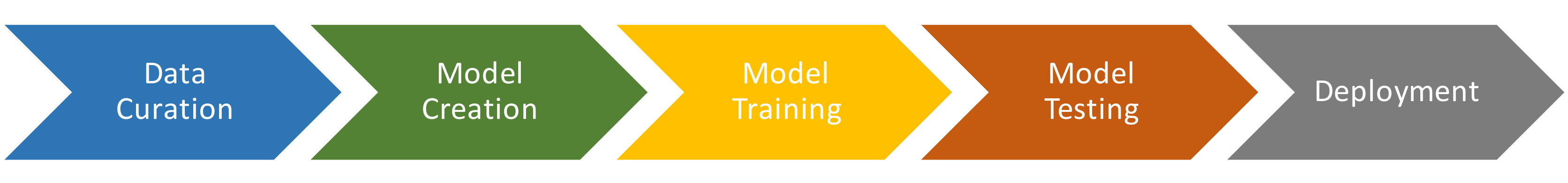}
\caption{Visualization of the standard machine learning pipeline.}\label{fig:ml_pipeline}
\end{figure}

The standard machine learning pipeline is mostly sequential by nature and its main steps are: 
\begin{enumerate}
\item {\textbf{Data curation}}, in which the dataset for training the model is created, encompassing the following phases: $(i)$ data collection, $(ii)$ data pre-processing, $(iii)$ data augmentation, and $(iv)$ data quality evaluation;
\item  \textbf{Model Creation}, in which the model is designed and built; 
\item \textbf{Model Training}, in which the created model is trained and in which the right hyperparameters are chosen (often through testing over a validation dataset); 
\item \textbf{Model Testing}, in which the performance of the model are assessed (ideally using different metrics and test sets). %
\item \textbf{Model Deployment}, in which the model is actually deployed in the real world.
\end{enumerate}
A visual representation of the standard sequential pipeline is given in Figure~\ref{fig:ml_pipeline}. Even though we do not show them in the figure, in practice loops-back might be necessary in order to modify the dataset or, way more often, the model. Still, all such possible changes happen mostly between the model creation step and the model training step, as they are usually driven by the desire to get better performance, while the dataset is often considered a static element of the pipeline. This is somehow not surprising given that often the unique requirement considered for machine learning models is to have good performance, according to some selected metrics.
However, as highlighted in the introduction, achieving a high score it is often not enough to ensure that the model will exhibit the right behaviour when used in practice. Other requirements may be put in place in order to make the model applicable or even certified to be applicable in the given application domain.

We now focus on two application domains---healthcare and autonomous driving---and we show how requirements arise naturally in both applications domains, how applying machine learning techniques has already brought and can bring tremendous advantages to the fields, and how deploying machine learning models in these fields without explicitly taking into account the requirements can lead to unexpected behaviors corresponding to violations of the requirements. Though we consider just two application domains, we believe that analogous considerations and results hold virtually in any application domain, given $(i)$ the body of knowledge developed along the years in any application domain, which can be translated in corresponding requirements, $(ii)$ the impressive results in performance obtained by machine learning models which will continue to push for their adoption despite the possible unintended behaviors, and $(iii)$ the impossibility to certify/rule out the absence of undesired behaviors without explicitly spelling out the corresponding requirements 
and the testing/verification of the model against the requirements themselves.

\subsection{Autonomous Driving}

In the recent years, the developments in the machine learning field, and in particular of the computer vision field, have fuelled the hopes of autonomous vehicles being finally in reach. However, every so often such hopes are crushed by car crashes that, in some cases, have even injured or even killed people. For example, in March 2018, an autonomous vehicle developed and tested on public roads by Uber's Advanced Technologies Group fatally injured a pedestrian who was pushing their bicycle across the street outside of a designated crossing area \cite{ubercrash}. The most striking characteristic of this accident is the fact that the car did not make any attempt to break and/or to avoid the pedestrian~\cite{ubercrash_journal}. Unfortunately, this is not an isolated incident, as nearly 400 car crashes involving autonomous vehicles have been reported in the United States over a period of only ten months in 2022~\cite{car-crashes-new-york-times}. 
As reported in~\cite{car-crashes-new-york-times}, these vehicles in order to take their decisions rely, among others, on computer vision models, which if trained to simply maximise their performance---following the standard machine learning development pipeline---might fail to abide to even the simplest requirements.

\begin{figure}[t]
    \centering
    \begin{subfigure}[b]{0.285\textwidth}
    \includegraphics[width=\linewidth]{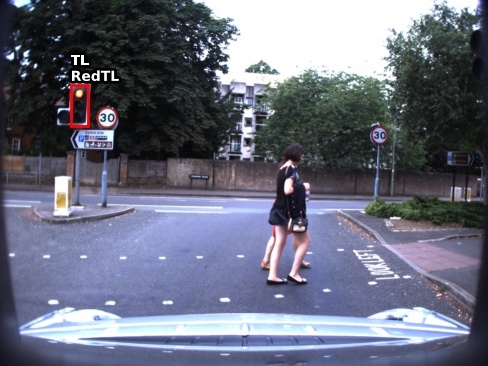}
    \caption{}
    \label{fig:autonomous_driving_ok}
    \end{subfigure}
    \hfill
    \begin{subfigure}[b]{0.285\textwidth}
    \includegraphics[width=\linewidth]{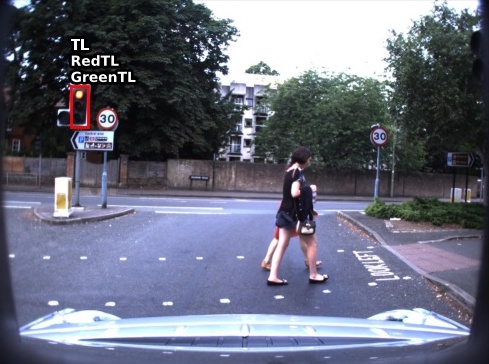}
    \caption{}
    \label{fig:autonomous_driving_viol}
    \end{subfigure}
    \hfill
        \begin{subfigure}[b]{0.40\textwidth}
    \includegraphics[width=\linewidth]{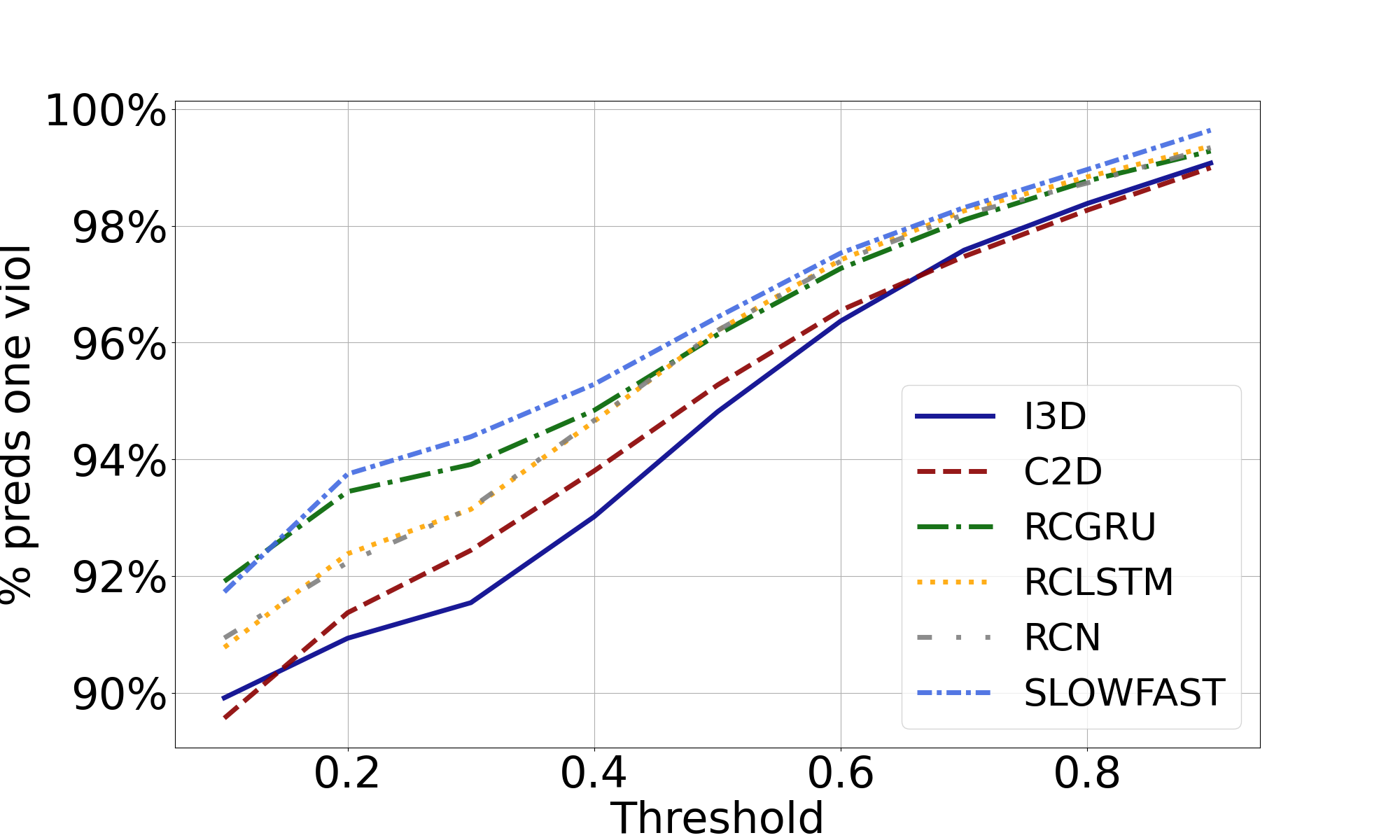}
    \caption{}
    \label{fig:num_violations}
    \end{subfigure}
\caption{Figure~\ref{fig:autonomous_driving_ok} and~\ref{fig:autonomous_driving_viol} show the predictions made by the I3D model (with $\theta=0.5$) for the same traffic light and just one frame apart. Figure~\ref{fig:num_violations} shows the number of predictions that violate at least one requirements when varying $\theta$. Figure~\ref{fig:num_violations} is from~\cite{giunchiglia2022_road}.}
\end{figure}

To exemplify the problem described above, we consider the ROAD-R dataset \cite{giunchiglia2022_road}, which consists of $(i)$ 22 relatively long ($\sim$8 minutes each) videos annotated with road events, i.e., a sequence of frame-wise bounding boxes linked in time, each labelled with the agent in the bounding box, together with its action(s) and location(s), and $(ii)$ 243 requirements expressed in propositional logic stating what is an admissible road event.\footnote{Dataset available at: https://github.com/gurkirt/road-dataset. Requirements available at: https://github.com/EGiunchiglia/ROAD-R.}  The possible labels associated to each bounding box are 41, and the requirements state which combinations of labels a model can output. Hence, for instance, a requirement states that a traffic light cannot be red and green at the same, while another states that a traffic light cannot move. Given ROAD-R, six state-of-the-art temporal feature learning architectures (I3D~\cite{carreira2017quo}, C2D~\cite{wang2018non}, RCGRU~\cite{hua2018traffic}, RCLSTM~\cite{hua2018traffic}, RCN~\cite{singh2019recurrent} and SlowFast~\cite{feichtenhofer2019slowfast}) as part of a 3D-RetinaNet model \cite{singh2021} (with a 2D-ConvNet backbone made of Resnet50 \cite{he2016}) for event detection, have been trained. These models take as input a set of consecutive frames, and for each frame they output $(i)$ a set of bounding boxes, and $(ii)$ a set of labels for each bounding box. Such labels are decided in the standard way: for each of the 41 labels the model outputs a number $o \in [0,1]$, and if $o > \theta$ then the label is returned, otherwise it is not. $\theta$ is a user-defined threshold which is often picked equal to $0.5$. An example of prediction is given in Figure~\ref{fig:autonomous_driving_ok}, where the prediction for the depicted bounding box is $\{\text{Traffic Light}, \text{Red Traffic Light}\}$. The surprising finding of the work~\cite{giunchiglia2022_road} is that, as shown in Figure~\ref{fig:num_violations}, no matter the chosen threshold $\theta$, at least 89\% of the predictions violate at least one requirement. Even more, some of the predictions may violate requirements corresponding to common knowledge possessed by humans, making the prediction difficult to interpret and manage. For example, if we consider the  prediction in Figure~\ref{fig:autonomous_driving_viol}, done by the same model which made the prediction in Figure~\ref{fig:autonomous_driving_ok}, for the same traffic light just one frame later, then we see that not only the prediction is wrong, but it also violates the common-knowledge (in this case corresponding also to a formally stated requirement) that the traffic light cannot be red and green at the same time. Such prediction, if not further elaborated by the system controlling the vehicle could indeed have dramatic consequences. For this reason, it is clear that any such prediction will have to be specifically handled by the system controlling the vehicle which for sure will incorporate and satisfy the violated requirements. However, if the requirements are incorporated into and surely verified by the entire system (as it must be the case in this setting), it is far from being clear why they are standardly neglected in the process of building the machine learning component of the system. Indeed, none of the six evaluated systems is able to handle requirements on their inputs-outpurs. Even more, as reported in \cite{giunchiglia2022_road}, the dataset used for training the model violates some of the constraints---like the fact that it is not possible for a vehicle to be both incoming and outgoing---which surely will have to be incorporated into any real application of the models.%
\footnote{While it might be argued that it is normal to have errors in the dataset because of noise, it seems odd to train a model with known-to-be-wrong-data and then correct the model outputs when they respect the known-to-be-wrong-data.}

\begin{figure}[t]
    \centering
    \begin{subfigure}[b]{0.48\textwidth}
    \includegraphics[width=\linewidth]{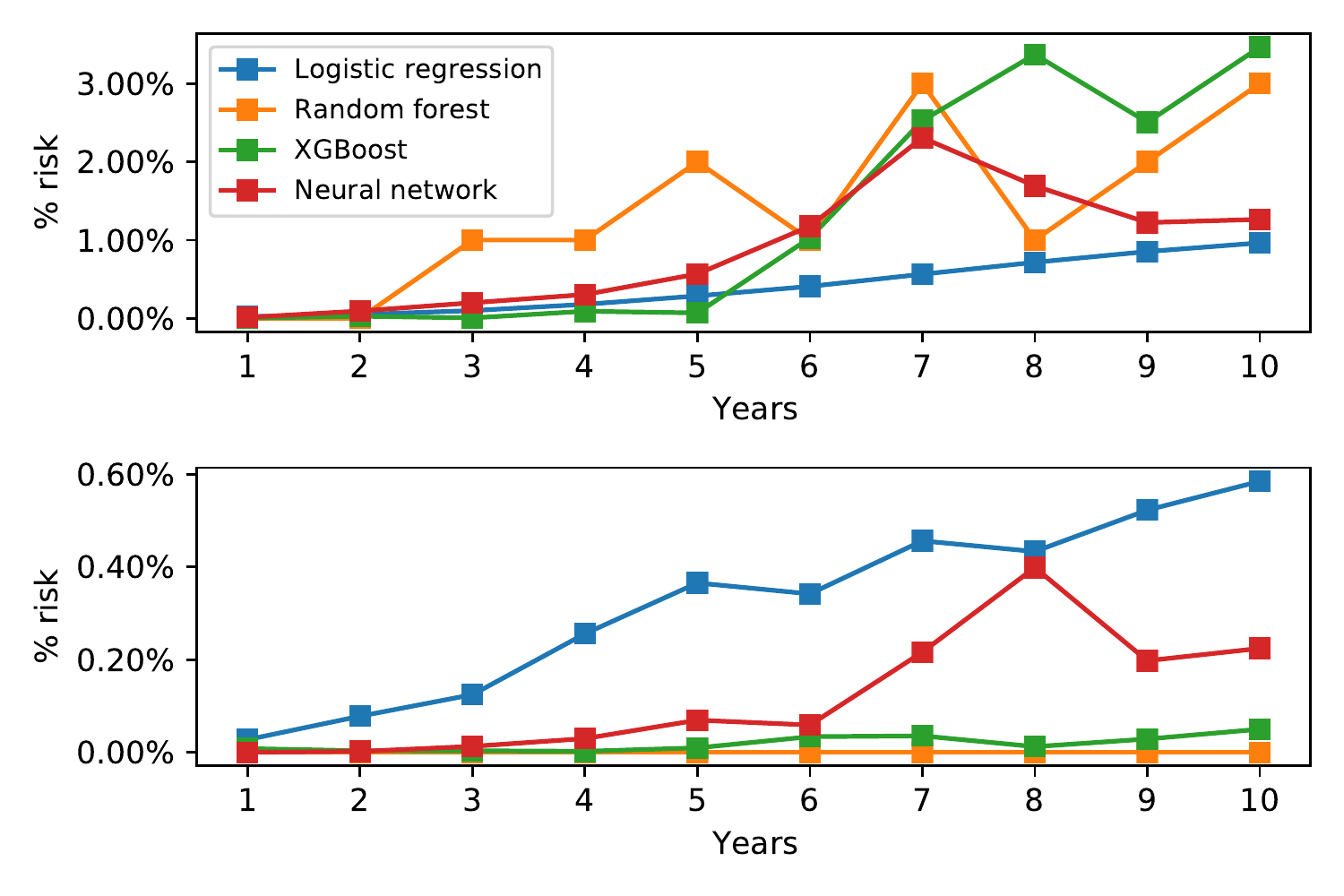}
    \caption{Example predictions at different time horizons.}
    \label{fig:healthcare_trajectory}
    \end{subfigure}
    \hfill
    \begin{subfigure}[b]{0.48\textwidth}
    \includegraphics[width=\linewidth]{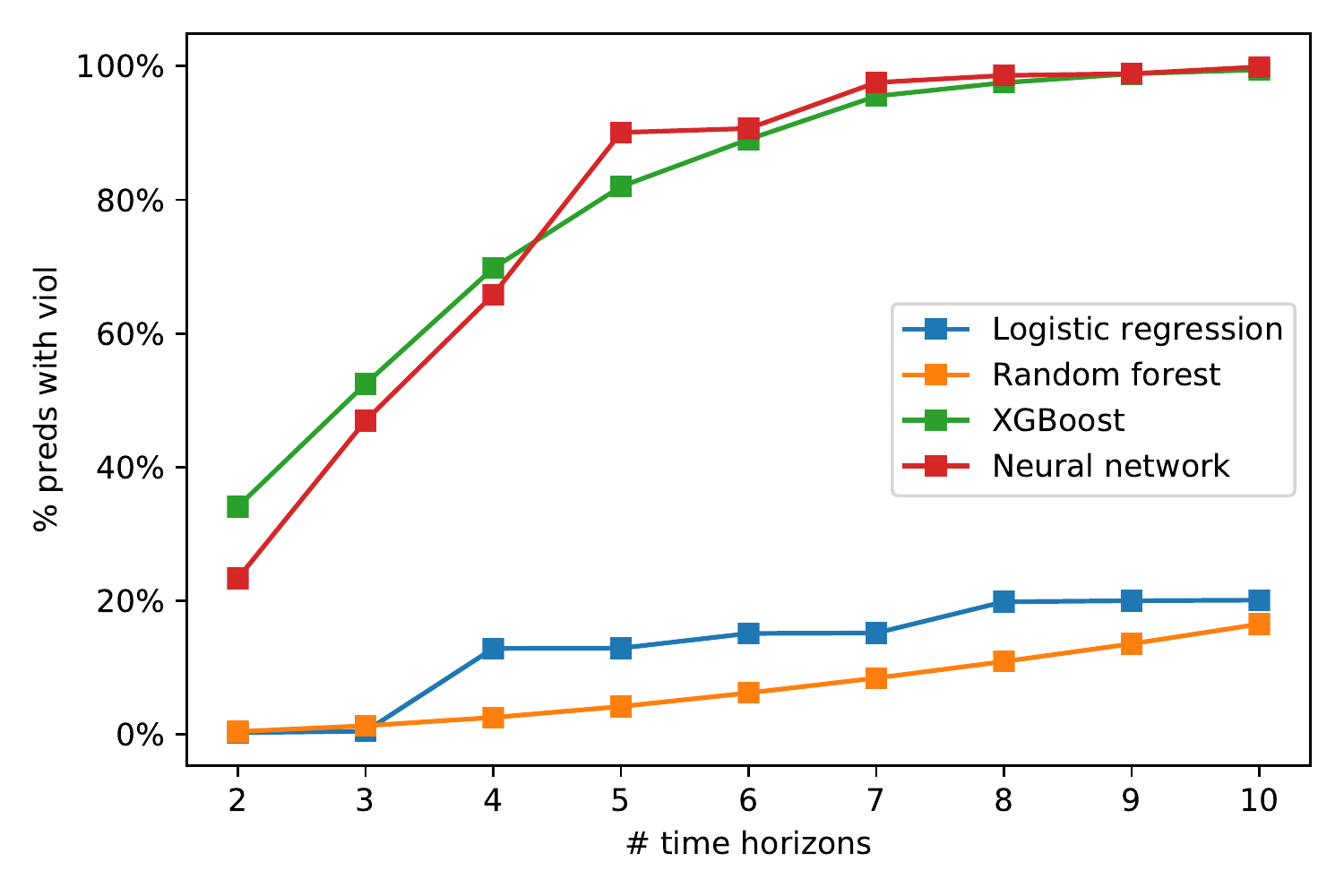}
    \caption{Proportion of patients with $\geq 1$ violation.}
    \label{fig:healthcare_viols}
    \end{subfigure}
\caption{Figure~\ref{fig:healthcare_trajectory} shows two examples of the predictions made by different classification models for the same patient at different time horizons. Figure~\ref{fig:healthcare_viols} shows the proportion of samples for which the predictions violate the requirement for risk to be increasing for longer time horizons.}
\label{fig:healthcare_eg}
\end{figure}

\subsection{Healthcare}

For our second example, we discuss healthcare.
There is great hope that developments in machine learning will revolutionize medicine and transform clinical practice \cite{Topol2019}.
The range of applications in medicine is vast, from computer vision systems analyzing medical images in radiology and longitudinal monitoring of patient trajectories throughout a hospital stay, to genomic screening of future disease risk and much more. 
However, as with many other high-stakes and safety-critical applications of machine learning, there are a number of requirements for machine learning systems in healthcare. These include ethical considerations, such as fairness and bias; practical considerations, such as controlling false positive rate to prevent alarm fatigue \cite{sendelbach2013alarm} and ensuring appropriate resource allocation; and logical considerations, such as the example discussed below, among others. 

As a concrete example, we consider clinical risk scores. Clinical risk scores estimate the likelihood of a specific outcome occurring after a certain period of time, such as a patient developing a particular disease or condition in the next ten years or experiencing an adverse event following a medical procedure.
By definition, the chance of an outcome occurring within some time horizon $t$ must increase with time, thus risk scores must be monotonically increasing functions of time, i.e., the predicted $n$-year risk is greater than their $m$-year risk, for all $n > m$.
However, there are many studies that use classification models to predict risk after a specific time horizon, for example 5-year risk. Naturally, we can generalize this by predicting risk at multiple time horizons rather than just one. Often this is very useful clinically since patient trajectories can vary greatly and patients with equivalent long-term risks might have divergent short-term risks.
A straightforward way to achieve this is to train multiple classification models to predict the risk at each time point of interest. However, there is no guarantee that the predicted risk will increase for longer time horizons. 

To illustrate the consequence of not incorporating the simple requirement described above into the machine learning systems, we consider the problem of predicting the risk of developing diabetes. 
We construct a cohort of patients from UK Biobank \cite{Sudlow2015}, a large-scale observational study with around 500,000 participants from 22 assessment centers across England, Wales, and Scotland enrolled between 2006 and 2010. 
We extracted a cohort of participants who were 40 years of age or older at enrollment with no diagnosis or history of diabetes at baseline. We considered the 18 features employed by QDiabetes \cite{HippisleyCox2017Diabetes}.
We performed data imputation using HyperImpute \cite{Jarrett2022} and trained classification algorithms using AutoPrognosis \cite{imrie2022autoprognosis} to predict $n$-year risk of developing diabetes for $n \in \{1, \dots, 10\}$. We averaged results over five random initializations.
Two examples are provided in Figure~\ref{fig:healthcare_trajectory}. For the first patient (Figure~\ref{fig:healthcare_trajectory}, top), only the logistic regression models met the requirement that the risk of developing diabetes should be monotonically increasing over longer time horizons, with the random forest, XGBoost, and neural network models having at least one time horizon with lower predicted risk than the previous. A similar situation can be seen for the second patient (Figure~\ref{fig:healthcare_trajectory}, bottom), where only the random forest model satisfies the requirement.
Considering the entire dataset and all ten time horizons, around 20\% of patients have predictions that violate the requirement for risk to monotonically increase at least once for the logistic regression and random forest models, while for over 99\% of patients XGBoost and neural networks issued predictions that violated the requirement.
Note that the number of violations made by the models was not necessarily correlated with performance, with the lowest-performing model as measured by area under the receiver operating curve at the 5-year horizon exhibiting the fewest violations (random forests).
Such predictions, at best, are simply inaccurate but, at worst, can erode trust in machine learning systems and have more serious consequences depending on the actions taken. 
There are several solutions to satisfy this particular requirement, such as specialized functional forms \cite{Cox1972,lee2018deephit}, multi-task learning with specialized loss functions \cite{li2016multi}, or bespoke architectures.
While the direct real-world impact of such predictions is likely limited for this particular example, the issue and need for requirements is clear.

\section{The Pyramid Model}
\label{sec:pyramid}

In the previous section we have shown the standard performance-driven machine learning development pipeline, in which the requirements are traditionally neglected, and, with the means of two examples, we have highlighted how such negligence can have dire consequences. This automatically calls for the inclusion of the requirements in the development process, which are at the core of any development pipeline proposed in the field of software engineering. We thus propose to adapt the general methodologies used
in software engineering for generic system development, where it is well known that the later the requirement are
taken into account the higher the cost is to repair the system.
In order to adapt such methodologies, we need to take into account that, differently from standard software, machine learning models are learnt from data, and thus we have lower control over the behaviour of the model, which entails not only that at design time it is possible to predict the future behaviour of the model, but also, at testing time, if a bug is found, then the steps necessary to fix it are entirely different.
A development pipeline for a machine learning model thus needs to take into account that: 
\begin{enumerate}
\item data are central to the process, as the quality of the model heavily depends on the quality of the collected data, 
\item the model is learnt from data, and thus it is not possible to fully know a priori its behaviour, and
\item if the model does not behave as expected, it is advisable to not only check the model but also check the data. 
\end{enumerate}
Further, it is unrealistic to expect that all requirements are always known a priori and that they remain immutable throughout the development process. This is already known in the general software engineering field, and it has been already specifically tested to be the case also in our field. Indeed, as highlighted by the survey conducted in~\cite{makinen2021mlops2}, some of the biggest problems that machine learning practitioners face on a daily basis include the accessibility of data---which might not be known in the requirements definition phase---and the unrealistic expectations of the stakeholders of the system---which might reveal themselves during the testing phase of the model. 
Thus, we also need to take into account that the requirements might change at each stage of the development process. %

\begin{figure}[t]
    \centering
    \qquad
    \includegraphics[width=0.6\linewidth]{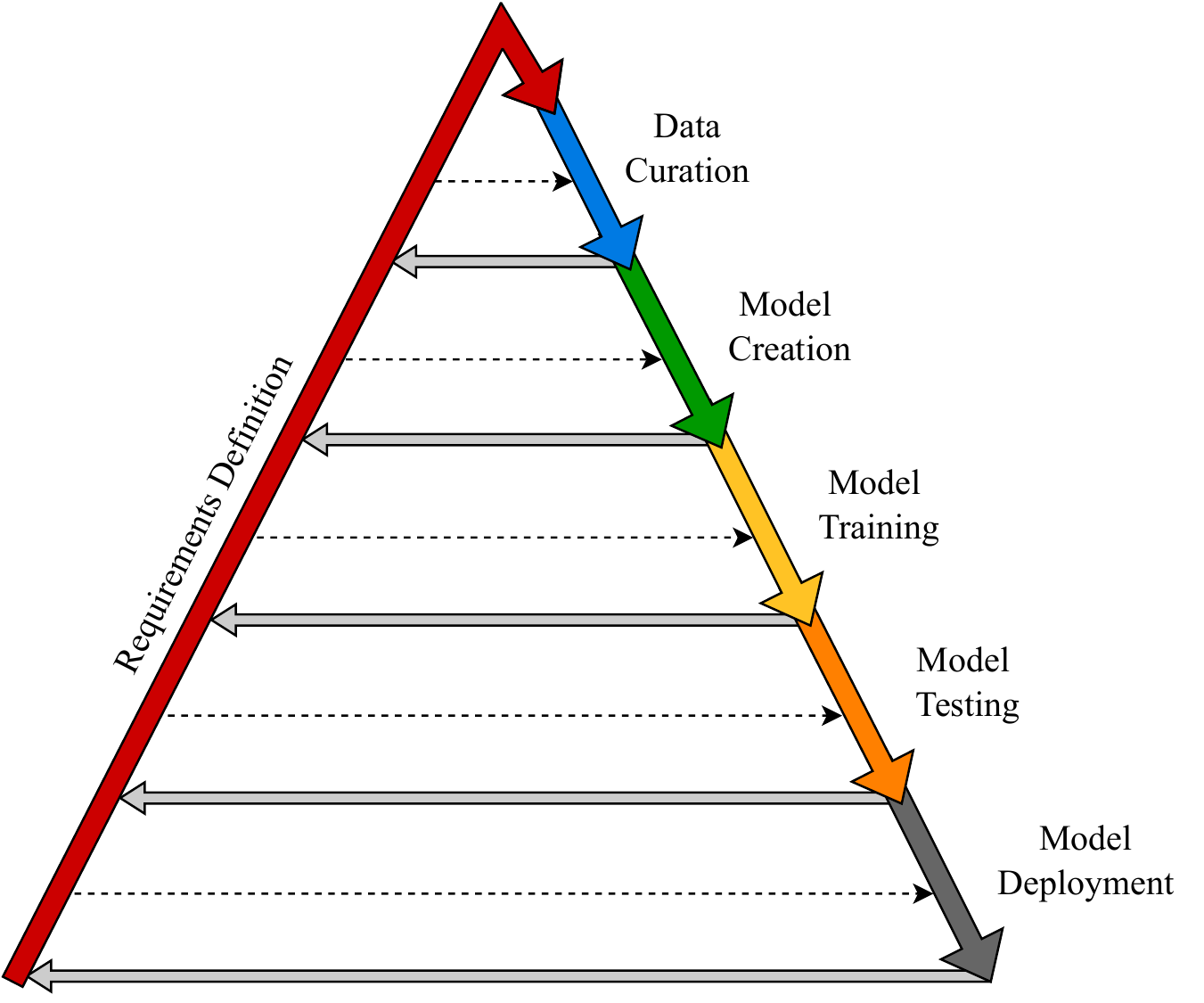}
    \caption{Visualization of the pyramid model. The full arrows stand for the standard procedural processes, while the dotted arrows show that the requirements impact every stage of the process.}
    \label{fig:pyramid}
\end{figure}

Given the above, we propose the  {\sl pyramid machine learning development model} in Figure~\ref{fig:pyramid}, in which 
the requirement definition phase is---as expected---put at the beginning of the development process
and in which the pyramid shape allows to better highlight the close relationship that must exist between the requirement definition phase and all the others in the pipeline.
Consider the grey backward lines. These represent the fact that if something unexpected happens at any point of the pipeline (e.g., the model does have some unintended behavior during the testing phase), then it is necessary to go back  and first check if the defined requirements were suitable or need to be updated, then if the collected data are representative of the phenomenon we want to capture, and so on. Obviously, the later in the pipeline we realise that there is a bug, the more costly it is, and this is captured in Figure~\ref{fig:pyramid} by the length of the path to cover as we advance in the pipeline.  Consider now  the dotted arrows, which illustrate the fact that the requirements can help in shaping each of the development steps, and in particular:
\begin{enumerate}
\item \textbf{Data Curation}: the requirements over the data obviously define the data collection phase, as they state which properties the dataset should have. This can positively impact this phase as it forces the stakeholders of the system to carefully reflect on which aspects of the total population the dataset should capture and which should be ignored. Further, they can help in reducing the mistakes done in the annotation phase---as any annotation that is not compliant with the requirements has to corrected---and they can even fasten it---as the annotators can be shown only those options that are compliant with the requirements or they can be automatically corrected in real-time.
\item \textbf{Model Creation}: the requirements over the model define how the model should behave, and thus they should be taken into account at creation time. Following the same principles used for the data curation step, having requirements can positively impact this phase because it forces the stakeholders of the system to reflect upon its expected behaviour and which outcomes must be avoided at all times.
Such requirements can then be mapped in the model itself. 
As an example, given a set of formal requirements, some works \cite{giunchiglia2021,hoernle2022multiplexnet,ahmed2022spl} in the neuro-symbolic field were able to map the requirements directly in the topology of neural networks and guarantee that the constraints are always satisfied, while exploiting the background knowledge expressed by the requirements to even get better performance. 
\item \textbf{Model Training}: the requirements over the model can again be used to define the objectives of our training. This is what it is done for example in many works in the fairness field, where the model is trained with two objectives: one to maximise performance and one to maximise the desired fairness definition (see, e.g., \cite{agarwal2018_fairness_loss}).  Further, if we map the requirements in the loss function, we can even use them  to alleviate the need for labelled data in semi-supervised settings (see, e.g., \cite{xu2018semanticloss,diligenti2012SBR,serafini2022ltn}).
Indeed, given an unlabelled datapoint, the model can be taught to simply return an output which is compliant with the requirements. 
\item \textbf{Model Testing}: as done in standard software engineering the requirements should guide the testing phase. Each requirement should be checked to hold (eventually using formal verification techniques, see, e.g., \cite{pulina2010,lomuscio2017}) and in case it does not, then either the requirement has to be further analyzed and eventually modified or
 some procedure has to be put in place in order to signal the requirement has to be properly handled outside of the machine learning model.
\item \textbf{Model Deployment}: requirements can shape also the deployment phase, especially if the stakeholders want the model to retain certain properties even in presence of data shift and/or drift.
\end{enumerate}

\section{Conclusions and the Road Ahead}\label{sec:concl}

In this paper we have shown that even though applying machine learning to a given problem can  bring great benefits, applying it without specifying and taking into account the requirements of the problem can lead to unexpected behaviors and, possibly, dramatic consequences. We have thus proposed a new machine learning development pipeline in which the requirements definition phase is explicitly incorporated at the beginning of the process, causing a deep connection between the requirements definition and all the other phases in the development process, in which---as standard in software engineering---the former may affect the latter and vice versa. From this perspective, this paper can be seen as a call to adopt the general methodologies used in software engineering for generic system development to the specific field of machine learning, highlighting the risks of not doing so.
We believe that in many cases the benefits of explicitly defining the requirements and taking them into account in the other phases outweigh their cost. It is also clear that the requirement definition and exploitation is  unavoidable when machine learning models are used as stand-alone systems in domains with strict requirements to satisfy (e.g., in safety critical domains) or when used as part of larger systems with a given set of stringent requirements (e.g., memory usage, see, e.g., \cite{bengio2016_memory}).
This view is supported by 
the recent trends in machine learning in which performance is just one of the requirements to satisfy.
In particular, 
the works done in the fields of interpretability (see, e.g., \cite{chen2019_interpretability,doshivelez2017_interpretability}), fairness (see, e.g., \cite{mehrabi2021fairness_survey,wick2019_fairness}) and robustness (see, e.g., \cite{weng2018iclr,katz2017_robustness}) all privilege certain characteristics of the model over performance, and all these works roots their motivation in the necessity to take into account requirements emerging from the respective application domain. However, how to 
define and incorporate the 
fully exploit the domain knowledge expressed by requirements in the development process is still largely an open question, as many technical challenges need to be overcome. To this end, the neuro-symbolic field can be seen as a bridge between standard machine learning and software engineering, as researchers in the field have already started developing techniques to create models which are built not only by the data, but which take explicitly  as input also other type of knowledge, like  formally stated requirements and use them either to design the model 
(see, e.g., \cite{giunchiglia2021,hoernle2022multiplexnet}), or to incorporate the requirements in the loss function (see, e.g., \cite{diligenti2012SBR,serafini2022ltn}).

Despite the huge amount of work that can be categorized under the umbrella of ``requirement driven'' machine learning, this is the first paper which explicitly advocates for a general model development methodology in which the requirement definition is a first class citizen like the other phases in the pipeline. We believe that this corresponds to a more structured approach to machine learning model development, which has already been advocated in \cite{gebru2018_datasheets_for_datasets,mitchell2019,mihaela2022-dc-check}.

\nocite{*}
\bibliographystyle{ios1}           %
\bibliography{bibliography}        %

\end{document}